\pdfoutput=1
\documentclass[11pt]{article}

\usepackage[]{ACL2023}

\usepackage{times}
\usepackage{latexsym}
\usepackage{graphicx,xcolor}  % グラフィックス関連
\usepackage{pxrubrica}        % ルビ
\usepackage{url}
\usepackage{booktabs}
\usepackage{multirow}
\usepackage{comment}

\usepackage[T1]{fontenc}
\usepackage[utf8]{inputenc}

\usepackage{microtype}

\usepackage{inconsolata}

\title{Towards Temporal Change Explanations from Bi-Temporal Satellite Images}

\author{Ryo Tsujimoto, Hiroki Ouchi, Hidetaka Kamigaito, Taro Watanabe \\
        Nara Institute of Science and Technology  \\
        \texttt{\{tsujimoto.ryo.tq0, hiroki.ouchi, kamigaito.h, taro\}@is.naist.jp}}

\begin{document}
\maketitle

\begin{comment}
%% 翻訳元
衛星画像から時系列変化を説明することは，都市計画や環境モニタリングなどにおいて重要である．
However, データセットの人手構築には高いコストがかかる。
In this paper, Large-scale Vision-Language Models (LVLMs) を利用した二時期衛星画像からの時系列変化説明手法を提案する。
我々は単一画像の入力にのみ対応したモデルで画像差分を説明させるため、All-at-OnceプロンプティングとStep-by-Stepプロンプティングの2種類のプロンプティングを考案した。
Finally, 時系列変化説明の精度に関して、プロンプティングおよびモデルを評価する。
我々の結果は、Step-by-Step プロンプティングの有効性、および、時系列変化説明の評価をするために必要な評価指標の策定の重要性を示す。
\end{comment}

\begin{abstract}
Explaining temporal changes between satellite images taken at different times is important for urban planning and environmental monitoring. 
However, manual dataset construction for the task is costly, so human-AI collaboration is promissing.
Toward the direction, in this paper, we investigate the ability of Large-scale Vision-Language Models (LVLMs) to explain temporal changes between satellite images. 
While LVLMs are known to generate good image captions, they receive only a single image as input.
To deal with a par of satellite images as input, we propose three prompting methods.
Through human evaluation, we found the effectiveness of our step-by-step reasoning based prompting.
%We employ models which can support only a single image input, and devised three types of prompting: The first method is called All-at-Once prompting, which involves concatenating bi-temporal SI side by side and generating the explanation in one go. The second method is called Step-by-Step prompting, which involves generating captions for each SI separately before generating the final explanation. The third prompting method is called Hybrid prompting, which combines these two prompting to leverage their respective strengths. Our results on remote sensing image change captioning dataset demonstrates the effectiveness of Step-by-Step prompting, but we have found the importance in establishing evaluation metrics necessary for assessing temporal change explanations.
\end{abstract}

\section{Introduction}

\begin{comment}
%%% 以下、翻訳元

衛星データは，空間分解能と波長分解能の両面で飛躍的な進歩を遂げ，利用者が入手することのできる衛星データの種類も多くなった．
中でも，撮影時期の異なる2枚の衛星画像（以下，bi-temporal SI）を用いた変化検出は，災害による被害状況の分析に用いられるなど，実応用上も重要な役割を果たしている．

bi-temporal SIからの変化検出として，従来はエッジ情報に基づいたハイライトや，マルチスペクタルデータに基づいた面積変化量の検出などが検討されてきた．
しかし，このような変化の検出だけではなく，その変化を言語化し，一般利用者にも直感的に理解可能な形式にすることが望ましい．
bi-temporal SIの時系列的変化説明の例を図1に示す．
ここでは，荒れ果てた並木道が住宅地に変わっていることが説明されている．

このような時系列変化説明生成に向けた既存データセットとして，ChenyangらのLevir-CCが存在する．
しかし，その説明は変化箇所を端的に描写した文となっている．

時系列変化説明として，変化箇所の描画だけでは不十分であり，変化の内容や程度における重要度を言語化することが望ましい．
たとえば，図1では，並木道から住宅地へのインフラ面での変化は，周辺の木々の増減といった自然環境の変化よりも重要度が高いと考えられる．そのため，変化の説明としてもこの点を強調した形で書かれることが望ましい．

この目的のために、我々の研究では、Large-scale Vision-Language Models (LVLMs)によってbi-temporal SIの時系列変化説明を生成する方法を調査する。

実験を行なうにあたり、単一画像の入力にのみ対応したモデルで時系列変化説明をさせるため、All-at-OnceプロンプティングとStep-by-Stepプロンプティングの2種類のプロンプティングを考案する。In addition, 生成された時系列変化説明の精度を検証するために、プロンプティングおよびモデルを複数の尺度で評価する。

Experimental result shows that，LLaVAにおいてAll-at-OnceプロンプティングよりもStep-by-Stepプロンプティングが有効であること。さらに、異なる評価指標として設定した忠実性と情報性の相関性も観察された。
\end{comment}
%The quality of satellite data has made remarkable progress in both spatial and spectral resolution, increasing the variety of satellite data available to users. Among these, change detection (CD) using two satellite images (SI) taken at different times (hereafter, \textbf{bi-temporal SI}) plays a crucial role in practical applications, such as analyzing damage caused by disasters~\cite{jaxa2021, nasa2021}.

Change detection (CD) using two satellite images (SIs) taken at different times (hereafter, bi-temporal SI) plays a crucial role in practical applications, such as analyzing damage caused by disasters~\cite{jaxa2021, nasa2021}.
Typical research on CD using bi-temporal SIs focuses on identifying pixel-level changes between images~\cite{AChangeDetectionMethod}, detecting and analyzing changes in geographical areas over time~\cite{ChangeDetectionMethods}, and using image differencing and ratioing techniques~\cite{DLMultispectral}. 

\begin{figure}[t]
    \centering
    \begin{tabular}{p{7cm}}
        \includegraphics[width=1\linewidth]{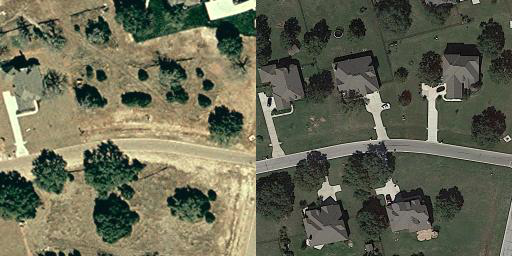} \\
%        The desolate tree-lined street has been transformed into a bustling residential neighborhood with houses and parked cars. \\
%        several buildings appear on both sides of the road.
%        some buildings are built beside the road with paths to the road.
%        several buildings have been constructed on the bareland on both sides of the road.
        Annotator A: \textit{``Five villas are built on both sides of the road.''} \\
        Annotator B: \textit{``Four buildings are built on both sides of the road.''}
    \end{tabular}
    \caption{Example of bi-temporal satellite images with their captions in Levir-CC; the left is the one before the change and the right is the one after the change.}
    \label{fig:example}
\end{figure}

Very recently, along with the advance of Large-scale Vision-Language Models (LVLMs),
a task of explaning changes between bi-temporal SIs has appeared.
An existing dataset for the task is Levir-CC\footnote{\url{https://github.com/Chen-Yang-Liu/RSICC}}~\cite{9934924}.\footnote{To the best of our knowledge, this is the only exsiting dataset.}
An example is shown in Figure~\ref{fig:example}, where the transformation of a dilapidated tree-lined street into a residential area is described.
%An example is shown in Figure~\ref{fig:example}.
One problematic issue is that explanations in this dataset are often too simple to comprehensively capture changes.
Another issue is that each bi-temporal SI pair has multiple captions with conflicting interpretations from different annotators.
In Figure~\ref{fig:example}, there is a conflict; e.g., ``Five villas are built'' and ``Four buildings are built.''
This example highlights the need for more comprehensive and consistent explanations.
Such explanations are costly for humans to create, so human-AI collaboration is promissing; e.g.,
LVLMs generate explanations and humans modify them.
Here, the cost of human modification depends on the performance of LVLMs; i.e., the better explanations LVLMs generate, the less modification humans do.

\begin{figure*}
    \centering
    \includegraphics[width=13cm]{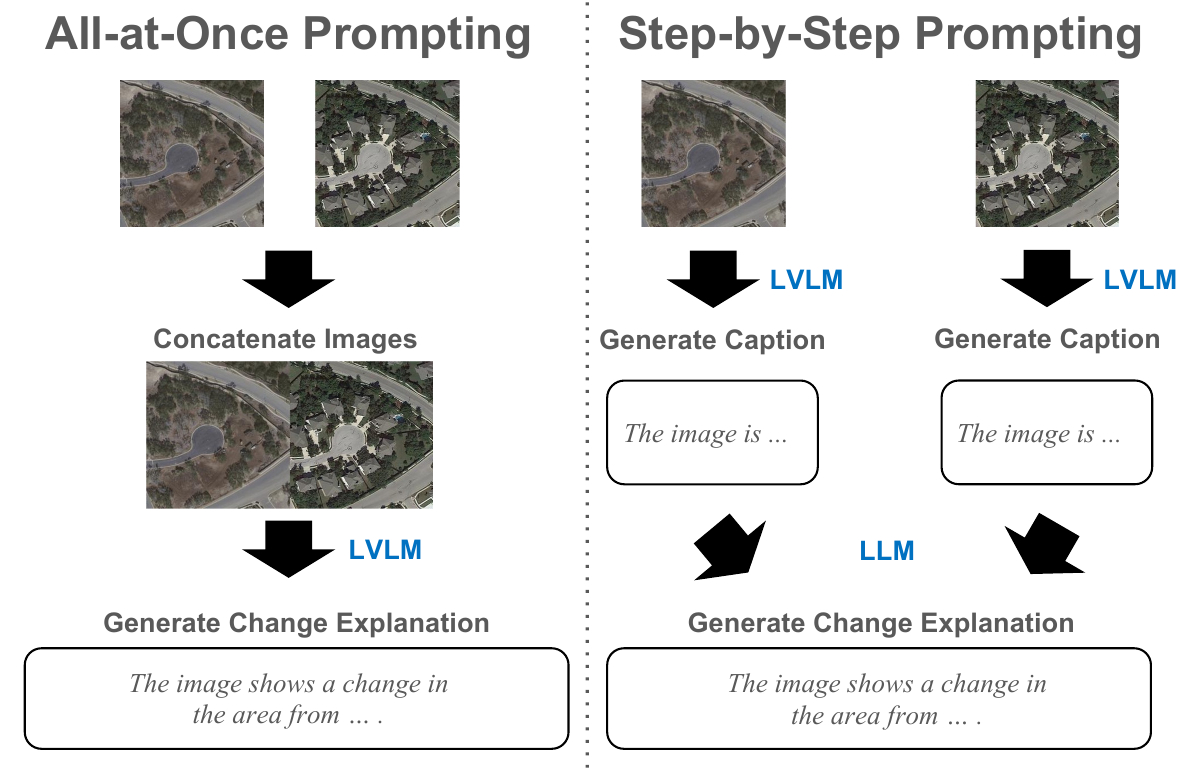}
    \caption{Explaining temporal changes from bi-temporal SI using two types of prompting}
    \label{fig:main}
\end{figure*}

Toward the human-LVLMs collaborative dataset construction, this study investigates the ability of LVLMs to explain temporal changes between bi-temporal SIs.
LVLMs are known for their ability to integrate visual information seamlessly with descriptive ability inherited from LLMs, allowing for nuanced explanations that align with image data~\cite{LVLM-eHub}.
In this paper, through human evaluation, we explore the efficacy of the explanations in meeting human expectations and understanding of temporal changes in visual data.
%To address these issues, our study investigates a method using Large-scale Vision-Language Models (LVLMs) to generate explanations for temporal changes from bi-temporal SI.
%LVLMs are known for their ability to integrate visual information seamlessly with descriptive capabilities inherited from LLMs, allowing for nuanced explanations that align with image data~\cite{LVLM-eHub}.
%Through the synthesis of LVLM-based explanations and human evaluations, we aim to explore the efficacy of these explanations in meeting human expectations and understanding of temporal changes in visual data.

The technical issue is that most LVLMs support only a single image as input, and thus, they cannot deal with two images. To address this issue, we propose three prompting methods: (i) All-at-Once, which involves concatenating bi-temporal SI side by side and generating explanations, (ii) Step-by-Step, which first generates captions for each SI separately and then generates the final explanation based on the captions, (iii) Hybrid, which combines these two promptings.
%In addition, to verify the accuracy of the generated temporal change explanations, we evaluate the prompting methods and models using multiple metrics.
To evaluate the methods, we define three metrics, Coverage, Truthfulness and Informativeness.
Through human evaluation in terms of the metrics, we found that Step-by-Step prompting with LLaVA-1.5\footnote{\url{https://github.com/haotian-liu/LLaVA}}~\cite{llava} is more effective than the All-at-Once prompting in all the metrics and All-at-Once prompting with GPT-4V~\cite{gpt_4v} achieves the best results in Truthfulness and Informativeness. 
%Furthermore, a correlation between trustfulness and informativeness, set as different evaluation metrics, was also observed.

\begin{comment}
%%% 翻訳元
LVLMsの中には、単一画像の入力にのみ対応しているものも存在する。
本手法では、図に示す2種類のPromptingと、それらを組み合わせたHybrid Promptingによる時系列変化説明生成を提案する．
具体的なプロンプト例は付録表1を参照のこと．

\subsection{All-at-Once Prompting}
このアプローチでは、bi-temporal SIを左右にconcatenationして単一画像としてLVLMに入力する。

\subsection{Step-by-Step Prompting}
LVLMsは、個別のSIに対するキャプショニング能力は高い。
しかし、事前実験によりAll-at-Onceプロンプティングによって生成される説明の精度の低さが判明した。
このアプローチでは、まずLVLMを用いて、時系列変化の前後の各SIについてキャプションを生成する。
そして，これらのキャプションをLLMに入力することで時系列変化を説明させる．
なお、各SIに関するキャプション生成の際には、より詳細な説明を期待して、Spatial Conceptsを含むようにキャプションに制約を与える。

\subsection{Hybrid Prompting}
All-at-OnceプロンプティングとStep-by-Stepプロンプティングを組み合わせたものである。
このアプローチでは、まずLVLMを用いて、各SIについてキャプションを生成する。
そして、これらのキャプションとbi-temporal SIを左右にconcatenationした画像を入力することで、時系列変化を説明させる。

\end{comment}

\section{Method}
This section describes our three prompting methods.
Actual prompt examples are provided in Table~\ref{tab:aao}, \ref{tab:sbs}, and \ref{tab:hybrid} of Appendix \ref{sec:appendix:Prompts}.

\paragraph{All-at-Once Prompting}
The bi-temporal SIs are concatenated side by side and input into the LVLM as a single image, shown in the lefthand in Figure~\ref{fig:main}.
Our intent behind this method is to allow the model to directly compare the two images and identify temporal changes in the single inference step.
%As shown in Figure~\ref{fig:main}, the bi-temporal SI is presented together.
%This allows the model to analyze the images placed next to each other simultaneously, facilitating a more straightforward identification of changes and enabling it to generate coherent explanations of these temporal changes.

\begin{table*}
    \centering
    {\small
    \begin{tabular}{cccccc} \toprule
        Prompting & Model & Coverage & Truthfulness & Informativeness & Avg num of words\\ \midrule
        All-at-Once & LLaVA-1.5 & 22.53 & 2.82 & 2.34 & 53.52\\
        Step-by-Step & LLaVA-1.5 $\rightarrow$ LLaVA-1.5 & 25.66 & 3.12 & 2.62 & 54.13\\
        Hybrid & LLaVA-1.5 $\rightarrow$ LLaVA-1.5 & \textbf{28.13} & 2.85 & 3.09 & 75.48\\ \midrule
        All-at-Once & GPT-4V & 16.95 & \textbf{3.56} & \textbf{3.24} & 62.78\\
        Step-by-Step &  LLaVA-1.5 $\rightarrow$ GPT-3.5-turbo & 22.22 & 3.01 & 1.58 & 58.46\\
        Step-by-Step & LLaVA-1.5 $\rightarrow$ GPT-4-turbo & 25.84 & 2.92 & 1.63 & 130.46\\
        Hybrid & GPT-4V $\rightarrow$ GPT-4V & 10.13 & 3.37 & 2.18 & 39.53\\
        \bottomrule
    \end{tabular}
    }
    \caption{Experimental results.}
    \label{tab:sample1}
\end{table*}

\paragraph{Step-by-Step Prompting}
LVLMs have a high capability for captioning individual SIs~\cite{rsgpt}.
The motivation behind this method is to address a potential issue with the All-at-Once prompting, where the model might not fully capture distinctive elements within the images.
In Step-by-Step prompting, captions for each SI before and after the temporal change are first generated using the LVLM.
Then, these captions are input into an LLM to explain the temporal changes.
To ensure more detailed explanations, captions for each SI are constrained to include spatial concepts such as places, spatial entities, and topological relations~\cite{pustejovsky-etal-2015-semeval}.

\paragraph{Hybrid Prompting}
This method integrates both All-at-Once prompting and Step-by-Step prompting.
The motivation behind this method is to address the limitations identified in each individual approach.
In All-at-Once prompting, the model might overlook detailed elements within the images because of the simultaneous presentation of both images.
Conversely, in Step-by-Step prompting, although detailed captions are generated, the model may miss the broader context of the change.
To mitigate these issues, captions for each SI are initially generated using the LVLM.
Then, the concatenated image of bi-temporal SIs and these captions are input to explain the temporal change, leveraging the strengths of both prompting methods to produce more comprehensive and accurate explanations.

\begin{comment}
%%% 以下、翻訳元
それぞれのモデルとプロンプティングを比較するため，様々な尺度を用いて評価実験を行った．

\subsection{実験設定}
\paragraph{モデル}
All-at-Once Promptingでは、concatenated bi-temporal SIを入力するLVLMsとして、LLaVA-1.5およびGPT-4-vision-preview(reffered to as GPT-4V)を利用した。
Step-by-Step Promptingでは、各SIからキャプションを生成するLVLMとしてLLaVA-1.5を利用した。また、時系列変化説明を生成するLLMとして、LLaVA-1.5、GPT-3.5-turboおよびGPT-4-turboを利用した。
Hybird Promptingでは、各キャプションの生成から時系列変化説明まで一括して、LLaVA-1.5もしくはGPT-4Vを利用した。

\paragraph{データセット}
bi-temporal SIと対応した時系列変化に関する5つの説明文を含むLevir-CCを用いた．Levir-CCは6815/1333/1929の訓練/検証/テスト用データを含み，半数は変化がないペアである．本実験では，テスト用データから抽出した100ペアのbi-temporal SIを用いて，各手法で生成された時系列変化説明を評価した．

\paragraph{プロンプトの制約}
文長を制限するために、in a clauseで説明がなされるようにinstructionで制約を与えた。

評価方法
自動評価
生成された説明に対して，Levir-CCにおける変化説明文中の名詞のcoverageを測る．
Levir-CCに含まれるGround TruthキャプションからspaCyライブラリで名詞を抽出し，生成された説明文に含まれる名詞の割合をcoverageとした．

人手評価
生成されたキャプションに対して，TruthfulnessとInformativenessを測る．
付録 表~\ref{tab:hitode}は各評価軸の実例である．
Truthfulness：Is it free of false information?。
忠実性1の例では「grass being replaced by a paved road」という誤りを含んでいる．
Informativeness：Does it describe detailed information or features of the image?。
情報性5の例では，「the transformation of the dirt road into a paved street」や「The trees in the left image have been replaced by houses」など，オブジェクトの位置や関係性についての記述を含む．
本実験では，英語話者のアノテーター2名に報酬を支払い、各評価軸について1から5までの5段階で回答してもらった．
\end{comment}

\section{Experiments}

\subsection{Experimental Setup}

\paragraph{Models}
In the All-at-Once prompting, we used LLaVA-1.5 and GPT-4V.
In the Step-by-Step prompting, we used LLaVA-1.5 as the LVLM to generate each SI caption. For generating the change explanation, we used LLaVA-1.5, GPT-3.5-turbo~\cite{gpt_3}, and GPT-4-turbo~\cite{gpt_4} as the LLMs.
In the Hybrid prompting, we used either LLaVA-1.5 or GPT-4V to handle the entire process from the caption generation to the change explanation.

\paragraph{Dataset}
We used Levir-CC, which includes five explanations for temporal changes of each bi-temporal SI pair.
%Levir-CC contains 6815/1333/1929 training/validation/test sets, with half being pairs without changes. 
We extracted 100 pairs of bi-temporal SIs from the test data of Levir-CC and evaluated the temporal change explanations generated by each method.

\paragraph{Prompt constraints}
To avoid reaching the token limit of LLMs, we instructed that explanations be limited to a single clause.

\subsection{Evaluation Metrics}
\label{metrics}

\paragraph{Automatic evaluation}
We measured the coverage of nouns in the generated explanations compared to the change explanation sentences in Levir-CC.
Nouns were extracted from the ground truth captions in Levir-CC using the spaCy library, and the proportion of these nouns present in the generated explanations was defined as ``Coverage.''

\paragraph{Manual evaluation}
We measured the Truthfulness and Informativeness~\cite{lin-etal-2022-truthfulqa} of the generated captions. Table~\ref{tab:hitode} of Appendix provides examples for each evaluation criterion.
\begin{description}
\setlength{\parskip}{0cm} 
\setlength{\itemsep}{0.1cm}
    \item \textbf{Truthfulness}: This criterion asks whether it is free from false information or not. For instance, an explanation with a truthfulness score of 1 in Table~\ref{tab:hitode} of Appendix includes the incorrect statement ``grass being replaced by a paved road.''
    \item \textbf{Informativeness}: This criterion judges whether it describes detailed information or features of the image or not. An example with an Informativeness score of 5 in Table~\ref{tab:hitode} of Appendix describes object positions and relationships, such as ``the transformation of the dirt road into a paved street'' or ``The trees in the left image have been replaced by houses.''
\end{description}

\noindent
We hired two English-speaking annotators to rate each criterion on a scale from 1 to 5.

\subsection{Experimental Results}
Table~\ref{tab:sample1} shows the results. 

\paragraph{Overall} For coverage, LLaVA-1.5 with Hybrid Prompting performed the best, indicating that this method is the most effective in identifying and describing all relevant changes in the images. 
In terms of Truthfulness and Informativeness, GPT-4V with All-at-Once Prompting was the best, suggesting that this approach provides the most accurate and detailed explanations. 

\paragraph{Comparison between prompting methods}
For LLaVA-1.5, we generated explanations using both All-at-Once Prompting and Step-by-Step Prompting. Step-by-Step Prompting outperformed All-at-Once Prompting across all evaluation metrics. This demonstrates the effectiveness of Step-by-Step Prompting. Additionally, the success of generating the final temporal change explanation using its own inferred information suggests that Chain of Thought Prompting~\cite{CoT} is also effective in visual tasks.

\subsection{Discussion}

\begin{figure}
    \centering
    \begin{tabular}{p{7cm}}
        \includegraphics[width=1\linewidth]{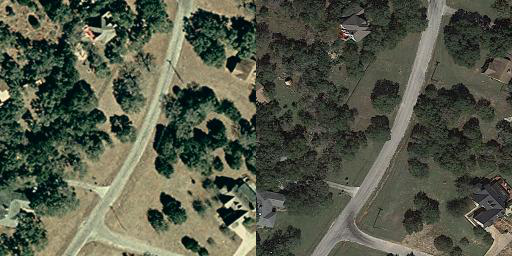} \\
        \textit{\small In comparing the left and right images of the same area, one can observe that a number of changes have taken place over time. In the right image, which represents the area after the changes, \textbf{additional roads have been constructed}, the landscape appears more developed with more houses built throughout the area, trees have been removed, and there is an overall increase in the infrastructure and human habitation, suggesting residential or suburban development.} \\
    \end{tabular}
    \caption{Example of Truthfulness score of 1}
    \label{fig:no_truthfulness}
\end{figure}

\begin{figure}
    \centering
    \begin{tabular}{p{7cm}}
        \includegraphics[width=1\linewidth]{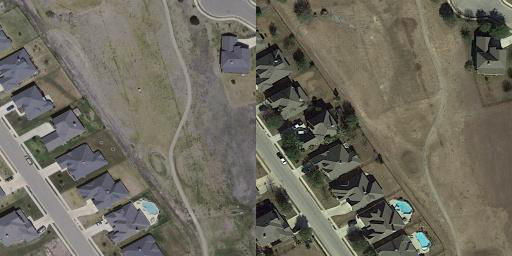} \\
        \textit{\small In the left image of the area before the change, the land appears largely undeveloped with clear lots and minimal construction activity, whereas the right image shows significant \textbf{development with completed houses, additional roads, and swimming pools}, indicating a transition from an undeveloped to a developed residential neighborhood.} \\
    \end{tabular}
    \caption{Example of Informativeness score of 1}
    \label{fig:no_informativeness}
\end{figure}

\paragraph{Analysis of low scores}
Figures~\ref{fig:no_truthfulness}~and~\ref{fig:no_informativeness} show examples of explanations rated 1 for Truthfulness and Informativeness, respectively. In the Truthfulness 1 case, the explanation included changes that did not exist. In the Informativeness 1 case, the explanation lacked specific details about the changes. Many low-scored outputs were from bi-temporal SIs with little to no change. This suggests that the models attempted to explain changes even when there were none, i.e., \textbf{over-explanation}. Therefore, future instructions should include exception handling for cases with no changes.

\paragraph{Relations between the metrics}
Sometimes, even if the Coverage score is high, the Informativeness score is low.\footnote{In the case of the example in Figure~\ref{fig:error} of Appendix, the Coverage is 62.5\% and yet Informativeness is rated 1.} 
%Despite containing a lot of information, this bi-temporal SI caption also includes the critical error, ``The black and white photograph.'' Moreover, there were many cases where Informativeness was rated low due to errors. 
There were many cases where Informativeness was rated low due to errors, while a lot of information is contained. 
The correlation coefficient between Truthfulness and Informativeness was 0.518, indicating a positive correlation. This suggests that humans are highly sensitive to errors. Even if an explanation contains rich vocabulary, a single mistake can significantly lower its Informativeness rating. This is because errors in the explanation greatly affect the overall trustworthiness and quality evaluation.
Additionally, concerning coverage, there were cases where coverage was high despite clear errors and low informativeness in the temporal change explanations. This is likely because the coverage of nouns in the ground truth is calculated, which includes commonly occurring words such as ``change,'' ``scene,'' and ``area.'' Therefore, preprocessing tailored to this task, such as setting stopwords and focusing on nouns directly related to observed changes in SIs, should be applied before calculating coverage to accurately reflect the relevance of the generated explanations to the ground truth.

\paragraph{Number of words in outputs}
The average number of words in explanations generated by the Step-by-Step prompting with GPT-3.5-turbo is 130.46 words. On the other hand, in the case of the Hybrid prompting with GPT-4V, the number is 39.53 words, which is approximately one-third of the former.
While the former outperforms the latter in both Truthfulness and Informativeness, the latter outperforms the former in Coverage.
%the latter is the coverage is significantly higher. 
As the explanation gets longer, the Coverage score tends to increase due to the larger number of possible covered words, while errors are more likely to be contained, which leads to lower scores of Truthfulness and Informativeness.
%Thus, there is a tendency for coverage to increase as the length of the explanation, even if erroneous, increases.

\begin{comment}
本研究では，LVLMsを用いて，bi-temporal SIの時系列変化説明を生成した．
また，評価実験により，Step-by-Step Promptingの有効性を確認した．
分析により，忠実性が低い場合には情報性も低い傾向にあり、これが情報性と網羅率の乖離の原因になっていることが明らかになった。
今後の展開としては、評価指標の再検討がある。SI Captioningタスクに合わせたトークン前処理の検討などが考えられる。
\end{comment}

\section{Conclusion}
\label{sec:contents-format4}

In this study, we used LVLMs to generate explanations for the temporal changes observed in bi-temporal SIs. Through the fair comparison using only Llama-1.5, we found the effectiveness of the Step-by-Step prompting. Also, All-at-Once prompting with GPT-4V is effective in terms of Truthfulness and Informativeness. 
Our analysis on low-scored outputs revealed the over-explanation problem; i.e., models attempt to explain changes even when there are none.
One future direction is remedying the over-explanation problem.
Another direction is to use the models built in this paper for constructing a comprehensive dataset for the task of explaining temporal changes observed in SIs.

%Our analysis revealed a tendency for informativeness to be low when truthfulness is also low, which has been identified as a contributing factor to the disparity between informativeness and coverage.Moving forward, there is a need for reassessment of evaluation metrics. This could involve considering token preprocessing tailored to the SI explanation task, among other potential adjustments.

\clearpage
\section*{Limitations}
We have not experimented with LVLMs capable of handling multiple images simultaneously. Therefore, our proposed methods are tailored to operate within the constraints of existing LVLM capabilities, potentially constraining their ability to effectively capture intricate temporal changes. Subsequent research should investigate the capabilities of multi-image LLMs to offer more comprehensive explanations of temporal changes and overcome the current limitations in the approach.

\bibliography{anthology,custom}
\bibliographystyle{acl_natbib}

\clearpage
\appendix
\section{Prompts}
\label{sec:appendix:Prompts}
%We generate temporal change explanations from bi-temporal SI by creating several prompts (see Table~\ref{tab:aao}, \ref{tab:sbs}, and \ref{tab:hybrid}).

%%% All-at-Once
\begin{table}[h]
    \centering
    \begin{tabular}{p{3cm}p{12cm}} \toprule
        Input for LVLMs &
        \begin{tabular}{l}
This image is a concatenation of two satellite images placed side by side.\\
- Both images show the same area.\\
- The left image shows the area before the change over time, while the right\\image shows it after.\\
Please describe where and what kind of changes in a clause, don't use \\bullet-points.\\ 
        \end{tabular}
        \\ \bottomrule
    \end{tabular}
    \caption{All-at-Once Prompting}
    \label{tab:aao}
\end{table}

%%% Step-by-Step
\begin{table}[h]
    \centering
    \begin{tabular}{p{3cm}p{12cm}} \toprule
    
        Input for LVLMs &
        \begin{tabular}{l}
Please provide a detailed description of the image.\\
The description should includes the following spatial concepts\\
- Places: toponyms, geographic and geopolitical regions, locations.\\
- Spatial Entities: entities participating in spatial relations.\\
- Paths: routes, lines, turns, arcs.\\
- Topological relations: in, connected, disconnected.\\
- Orientational relations: North, left, down, behind.\\
- Object properties: intrinsic orientation, dimensionality.\\
- Frames of reference: absolute, intrinsic, relative.\\
- Motion: tracking objects through space over time. 
        \end{tabular}
        \\ \midrule
        
        Input for LLMs&
        \begin{tabular}{l}
description of the area before the change: <DESCRIPTION>\\
description of the area after the change: <DESCRIPTION>\\
Please describe where and what kind of changes occurred in the area,\\in a clause.
        \end{tabular}
        \\ \bottomrule

    \end{tabular}
    \caption{Step-by-Step Prompting}
    \label{tab:sbs}
\end{table}

\clearpage
%%% Hybrid
\begin{table}[h]
    \centering
    \begin{tabular}{p{3cm}p{12cm}} \toprule
    
        Input for LVLMs&
        \begin{tabular}{l}
Please provide a detailed description of the image.\\
The description should includes the following spatial concepts\\
- Places: toponyms, geographic and geopolitical regions, locations.\\
- Spatial Entities: entities participating in spatial relations.\\
- Paths: routes, lines, turns, arcs.\\
- Topological relations: in, connected, disconnected.\\
- Orientational relations: North, left, down, behind.\\
- Object properties: intrinsic orientation, dimensionality.\\
- Frames of reference: absolute, intrinsic, relative.\\
- Motion: tracking objects through space over time. 
        \end{tabular}
        \\ \midrule
        
        Input for LVLMs&
        \begin{tabular}{l}
description of the area before the change: <DESCRIPTION>\\
description of the area after the change: <DESCRIPTION>\\
Please describe where and what kind of changes occurred in the area,\\ in a clause.
        \end{tabular}
        \\ \bottomrule

    \end{tabular}
    \caption{Hybrid Prompting}
    \label{tab:hybrid}
\end{table}
\section{Annotation}

%%% hitode
\begin{table*}[h]
    \centering
    \begin{tabular}{cccp{5cm}} \toprule
        Satellite Images & Metrics & Score & Example\\ \midrule
        \multirow{2}{*}{\includegraphics[width=5cm]{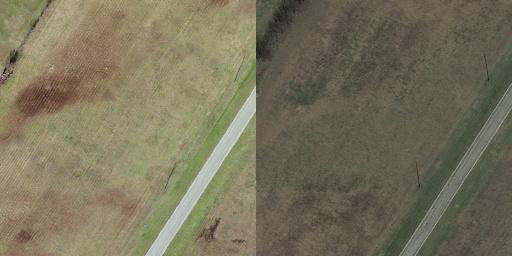}} & \multirow{2}{*}{Truthfulness} & 1 &
        \textit{The changes in the area include the grass being replaced by a paved road, which is visible in the right image.
        This transformation likely occurred due to urbanization or infrastructure development, leading to the conversion of \textbf{the grassy field into a paved road}.}\\
         &  & 5 & 
         \textit{In the image on the left, there is a stretch of land with healthy green vegetation on either side of a well-defined white road, and some brownish patches indicate areas where the vegetation is sparse or dead. By contrast, in the image on the right, the overall color of the terrain is duller.}\\
        \multirow{2}{*}{\includegraphics[width=5cm]{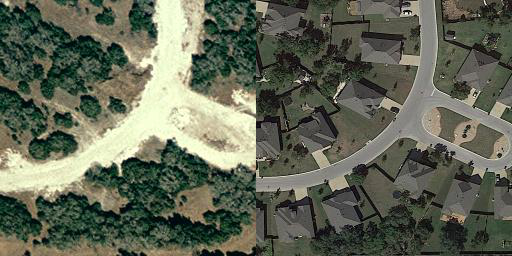}} & \multirow{2}{*}{Informativeness}  & 1 & 
        \textit{The change in the image is the construction of a new road or street, which is visible in the right-side image.}\\
         &  & 5 & 
         \textit{In the image, there is a noticeable change in the area, with the left image showing a dirt road surrounded by trees, while the right image displays a neighborhood with a street intersection and houses. The changes that have occurred include the transformation of \textbf{the dirt road into a paved street}, the addition of houses and street infrastructure, and the presence of multiple cars parked or driving through the intersection. \textbf{The trees in the left image have been replaced by houses}, and the overall appearance of the area has shifted from a rural setting to a more urban, residential neighborhood.}\\ \bottomrule
    \end{tabular}
    \caption{Manual evaluation}
    \label{tab:hitode}
\end{table*}

\clearpage
\section{Case Study}
\label{sec:appendix:CaseStudy}
\begin{figure}[h]
    \centering
    \begin{tabular}{p{15cm}}
        \includegraphics[width=1\linewidth]{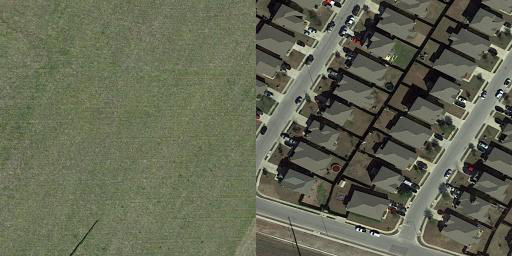} \\
        \textit{
        The described change in the area is significant, transforming from a natural, rural environment to a developed urban or suburban residential setting.
        }
        
        \textit{
        Before the change, the area was characterized by a large expanse of green space, occupied almost entirely by grass, with a single stick noticeable in the center of the field. This indicates a relatively untouched or undeveloped landscape that offers a tranquil and natural atmosphere, potentially used for agricultural purposes or as a natural habitat.
        }
        
        \textit{
        After the change, the area underwent substantial urban development. The introduction of multiple houses and cars signifies that the field has been replaced by a residential neighborhood. The houses are lined up in rows which suggest a planned community, and the presence of parked cars indicates the area is now inhabited and that residents likely rely on personal vehicles for transportation. \textbf{The black and white photograph} gives the impression that the image is either historical or that it's stylized to evoke a sense of the past. The closeness of the houses to one another points to a tightly-knit community layout, common in suburban developments.
        }
        
        \textit{
        In essence, the area has shifted from being an open, natural space to a built environment with residential housing, representing a change from undeveloped land to a structured neighborhood with human habitation and associated infrastructure. This kind of alteration typically involves construction, land clearing, the establishment of roads, and the implementation of utilities and services to support the residential community.
        } \\
    \end{tabular}
    \caption{Example output for informativeness score of 1}
    \label{fig:error}
\end{figure}

\clearpage
\section{Analysis}
\label{sec:appendix:analysis}
\begin{figure}[h]
    \centering
    \includegraphics[scale=1]{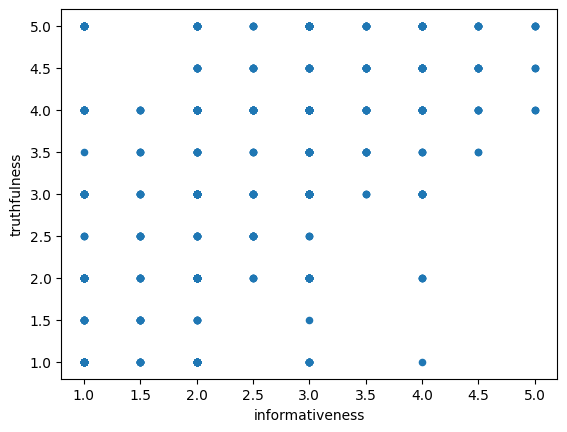}
    \caption{Scatter Plot of Informativeness and Truthfulness}
\end{figure}

\end{document}